\documentclass[11pt, a4paper, logo]{deepmind}

\usepackage{amsmath, latexsym}
\usepackage{amsfonts}
\usepackage{mathtools}
\usepackage{ntheorem}
\usepackage{dsfont}
\usepackage[dvipsnames]{xcolor}
\usepackage[colorinlistoftodos]{todonotes}
\usepackage{booktabs}
\usepackage{xfrac}
\usepackage{bbm}

\usepackage[most]{tcolorbox}
\usepackage{xparse}
\usepackage{lipsum}
\usepackage{changepage}
\usepackage{enumitem}

\newcommand{\squishlist}{
   \begin{list}{$\bullet$}
    { \setlength{\itemsep}{0pt}      \setlength{\parsep}{3pt}
      \setlength{\topsep}{3pt}       \setlength{\partopsep}{0pt}
      \setlength{\leftmargin}{1.5em} \setlength{\labelwidth}{1em}
      \setlength{\labelsep}{0.5em} } }

\newcommand{\squishlisttwo}{
   \begin{list}{$\bullet$}
    { \setlength{\itemsep}{0pt}    \setlength{\parsep}{0pt}
      \setlength{\topsep}{0pt}     \setlength{\partopsep}{0pt}
      \setlength{\leftmargin}{2em} \setlength{\labelwidth}{1.5em}
      \setlength{\labelsep}{0.5em} } }

\newcommand{\squishend}{
    \end{list}  }

\newcommand{\two}{(2)}

\DeclareMathAlphabet{\mathpzc}{OT1}{pzc}{m}{n}

\usepackage{hyperref}       
\usepackage{url}            
\usepackage{booktabs}       
\usepackage{amsfonts}       
\usepackage{nicefrac}       
\usepackage{microtype}      
\usepackage{balance}
\usepackage{amsmath,amssymb}
\usepackage{layouts}  
\usepackage{graphicx}

\usepackage{algorithm,algorithmic}
\makeatletter
\newcommand{\ALOOP}[1]{\ALC@it\algorithmicloop\ #1\begin{ALC@loop}}
\newcommand{\ENDALOOP}{\end{ALC@loop}\ALC@it\algorithmicendloop}

\makeatother

\DeclareMathOperator*{\EXP}{\mathbb{E}}
\DeclareMathOperator*{\ARGMAX}{\mathrm{argmax}}

\DeclareMathOperator{\pit}{\tilde{\pi}}
\DeclareMathOperator{\KL}{D_{KL}}
\DeclareMathOperator{\LOSS}{\mathfrak{L}}

\DeclareMathOperator*{\softmax}{\mathrm{softmax}}

\newcommand*\diff{\mathop{}\!\mathrm{d}}

\usepackage[numbers, sort&compress, square]{natbib} 

\title{Quinoa: a Q-function You Infer Normalized Over Actions}

\correspondingauthor{grave@deepmind.com}

\renewcommand{\today}{2018-12-08} 

\author[1]{Jonas Degrave}
\author[1]{Abbas Abdolmaleki}
\author[1]{Jost Tobias Springenberg}
\author[1]{Nicolas Heess}
\author[1]{Martin Riedmiller}

\affil[1]{DeepMind}

\begin{abstract}
We present an algorithm for learning an approximate action-value soft Q-function in the relative entropy regularised reinforcement learning setting, for which an optimal improved policy can be recovered in closed form. We use recent advances in normalising flows for parametrising the policy together with a learned value-function; and show how this combination can be used to implicitly represent Q-values of an arbitrary policy in continuous action space. Using simple temporal difference learning on the Q-values then leads to a unified objective for policy and value learning. We show how this approach considerably simplifies standard Actor-Critic off-policy algorithms, removing the need for a policy optimisation step. We perform experiments on a range of established reinforcement learning benchmarks, demonstrating that our approach allows for complex, multimodal policy distributions in continuous action spaces, while keeping the process of sampling from the policy both fast and exact.
\end{abstract}

\begin{document}
\maketitle
\balance

\section{Introduction}


Off-policy actor-critic algorithms, in combination with deep neural networks, hold promise for solving problems in continuous control, as they can be used to learn complex non-linear policies in a data-efficient manner~\cite{peters2008natural}. Typically deep actor-critic approaches consist of two steps. First, a neural network is used to fit the Q-values of the current policy. After that, a parametric policy -- often a conditional Gaussian distribution -- is learned by maximising these learned Q-values. These two steps are then iterated to convergence. Ideally, the second policy optimisation step would not be needed. After all, optimising a policy against a learned Q-function just transforms action-preferences into a normalised distribution. This optimisation step cannot produce new information which was not already encoded in the Q-function. It can however introduce sub-optimal behaviour through approximation errors; either due to the choice in the parametric policy distribution or due to numerical fitting errors. 

As a consequence, the idea of learning a Q-function from which an improved policy can be obtained without additional optimisation, has been previously considered by learning normalised advantage functions (NAF)~\cite{baird1994reinforcement,gu2016continuous} or using compatible function approximation with Gaussian policies~\cite{sutton2000policy}. While appealing in theory, these approaches come with the caveat that they put additional constraints on the Q-function, such as being locally quadratic in action space. This limits the expressiveness of the Q-function, making it no longer able to correctly fit any set of Q-values.

In this work, we propose an algorithm which learns a soft Q-function globally while providing the optimal policy in closed form. We call this algorithm Quinoa, a \emph{Q-function you Infer Normalised Over Actions} as we can directly perform inference on the optimal policy, which is the soft Q-function normalised over the action dimensions. We find that the key to allowing unrestricted Q-functions that allow for inference of the optimal policy, is to use a richer class of policy parametrisations. In particular, we use normalising flows as they can be universal density function approximators.  In the next section, we will explain how we derive our soft Q-function, starting from a relative entropy regularised RL objective, as considered in REPS~\cite{peters2010relative}, TRPO~\cite{schulman2015trust}, MPO~\cite{abdolmaleki2018maximum} and SAC~\cite{haarnoja2018soft}.

\section{Background}

We consider the standard discounted reinforcement learning (RL) problem defined by a Markov decision process (MDP). The MDP consists of continuous states $s$, actions $a$, transition probabilities $p(s_{t+1} | s_t, a_t)$ which specify the probability of transitioning from state $s_t$ to $s_{t+1}$ under action $a_{t}$, a reward function $r(s, a) \in \mathbb{R}$ and the discount factor $\gamma \in [0, 1)$. The policy $\pi_\theta(a | s)$ with parameters~$\theta$ is a probability distribution over actions $a$ given a state $s$. For brevity, we drop the subscript $\theta$ in the following. Together with the transition probabilities, these give rise to a state-visitation distribution $\mu_\pi(s)$. We consider the relative entropy regularised RL setting that encourages the policy to trade off reward with policy entropy. Unlike the regular expected reward objective, this can provide advantages when it is desirable to learn multiple solutions for a given task. More generally, it can help to regularise the policy by preventing collapse of the state-conditional action distribution. We define the relative entropy of policy $\pi$ compared to a reference policy $\pit$ as:
$
\KL[\pi,\pit|s] = \EXP_{a \sim \pi(\cdot | s)}\left[ \log\left(\frac{\pi(a | s)}{\pit(a | s)}\right) \right],
$
noting that if $\pit$ is uniform we recover the entropy $\mathcal{H}[\pi]$.
The goal for the RL algorithm is to maximise the expected sum of discounted future returns, regularised with this relative entropy:
\begin{equation*}
J(\pi) = \EXP_{\pi,p} \Bigg[ \sum_{i=0}^\infty \gamma^i r_t(s_i, a_i) - \alpha \KL[\pi,\pit|s_i]\Bigg| s_0, a_0, a_i \sim \pi(\cdot | s), s_i \sim p(\cdot | s_{i-1}, a_{i-1}) \Bigg].
\end{equation*}

We define the soft action-value function associated with policy $\pi$ as the expected cumulative discounted return when choosing action $a$ in state $s$ and acting subsequently according to policy $\pi$  factor $\gamma$, as 
$$
Q^s_\pi(a, s) = J(\pi)_{\lbrace s_0 = s, a_0 = a \rbrace}  = \EXP_{\pi, p} \left[ \sum_{i=0}^\infty \gamma^i r_t(s_i, a_i) - \alpha \KL[\pi,\pit|s_i] \Bigg| s_0=s, a_0=a \right].
$$
Observing that the action at time $t$ does not influence the KL at $t$, this action value function can be expressed recursively as $$
Q^s_\pi(a_t, s_t) = \EXP_{s_{t+1} \sim p(\cdot | s_t, a_t)} \left[ r(s_t, a_t) + \gamma V^s_\pi(s_{t+1}) \right],
$$ where $V^s_\pi(s) = \EXP_\pi[ Q^s_\pi(s,a) ] - \alpha \KL[\pi,\pit|s]$ is known as the soft value function of $\pi$. 

\begin{figure}[t]
\centering
\begin{minipage}[c]{1\textwidth}
\def\mywidth{0.495}
\centering
\includegraphics[width=\mywidth\textwidth]{figures/plot-cheetah.pdf}
\includegraphics[width=\mywidth\textwidth]{figures/plot-walker.pdf}
\includegraphics[width=\mywidth\textwidth]{figures/plot-hopper.pdf}
\raisebox{0.09\textwidth}{\includegraphics[width=0.16\textwidth]{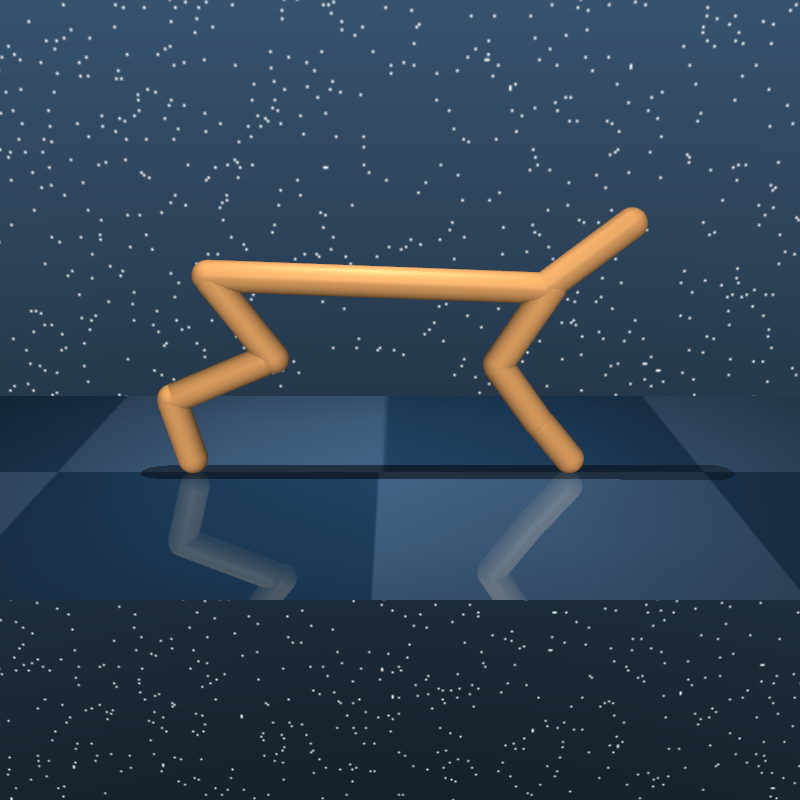}}
\raisebox{0.09\textwidth}{\includegraphics[width=0.16\textwidth]{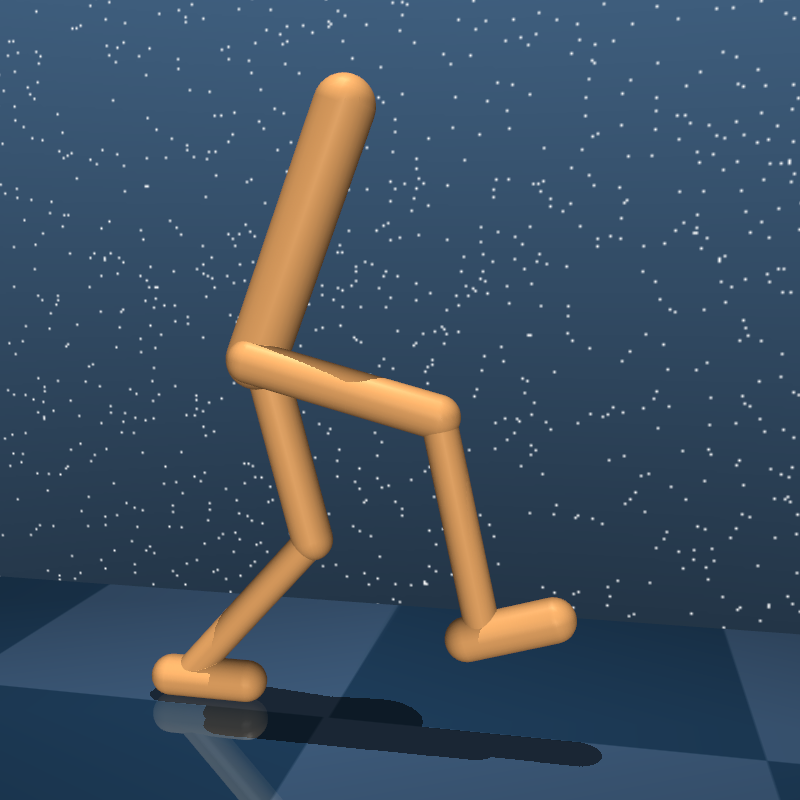}}
\raisebox{0.09\textwidth}{\includegraphics[width=0.16\textwidth]{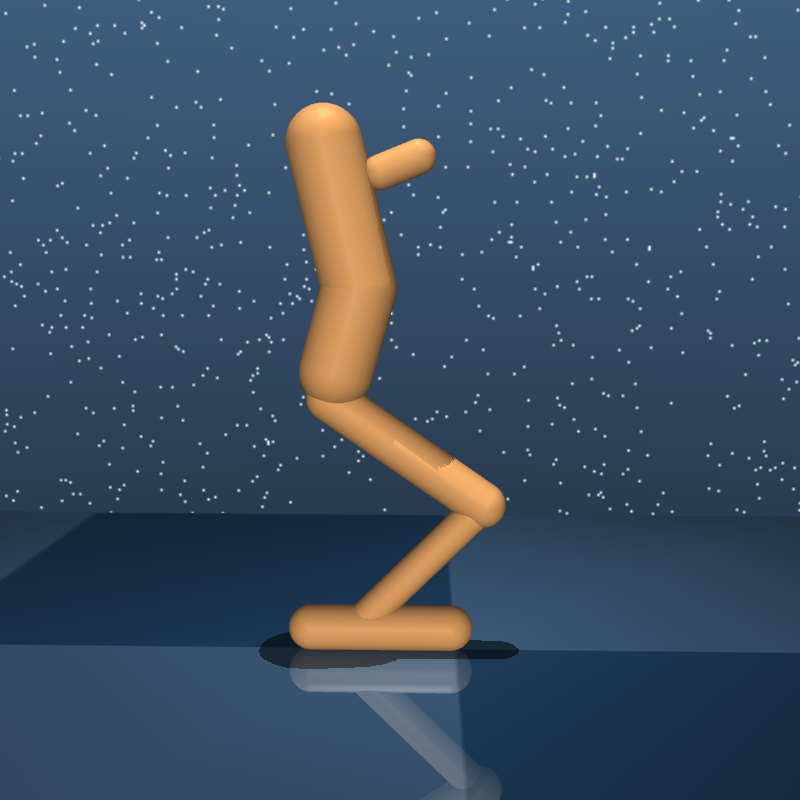}}
\caption{The performance of Quinoa compared with SVG(0)~\cite{heess2015learning}. Shown are the median performance across 5 seeds, together with the minimum and maximum performance. We show these performances for three domains from the DeepMind Control Suite as illustrated on the bottom right; from left to right: Cheetah, Walker and Hopper.}
\label{fig:charts}
\end{minipage}
\end{figure}

\section{Quinoa}

We would like to find a soft-optimal policy $\pi$, which in every state maximises the soft Q-function $Q^s_\pi(a, s)$. This in turn would locally maximise our objective $J(\pi)$ at each state under the assumption that $Q^s_\pi$ is sufficiently accurate. Solving for $\pi$ comes with one caveat: finding the multiplier $\alpha$ trading the regularisation with the reward is hard, as the magnitude of the reward can differ significantly over the course of the training process. We therefore optimise $Q^s$ subject to a hard constraint on the relative entropy between the policy $\pi$ and the prior $\pit$, as the parameters through that approach are easier to set in practice~\cite{abdolmaleki2018maximum,peters2010relative}:
\begin{equation*}
\pi = \ARGMAX_\pi \EXP_{s\sim\mu_\pi} \big[Q^s_\pi(a,s)\pi(a|s)\big] \enspace\textrm{subject to}\enspace \EXP_{s\sim\mu_\pi}\big[\KL[\pi,\pit|s]\big] < \epsilon\enspace\textrm{and}\enspace \forall s: \EXP_{a}\big[\pi(a|s)\big] = 1,
\end{equation*}
where the last constraint ensures that $\pi$ is normalised. We solve this constrained optimisation problem using the method of Lagrange multipliers, automatically obtaining an optimal $\alpha$ for a given $\epsilon$. The details of this procedure are given in the Appendix.

Taking into account the constraint that $\pi(a|s)$ is a distribution, we obtain that the optimal policy $\pi(a|s)$ for a given $Q^s_\pi(a,s)$ is given by 
\begin{equation}
\pi(a|s) = \frac{\pit(a|s) \exp\Big(\frac{Q^s_\pi(a, s)}{\alpha}\Big)}{\int \pit(a'|s) \exp\Big(\frac{Q^s_\pi(a', s)}{\alpha}\Big) \diff a'}
\label{eq:start}
\end{equation}

this not only defines an improved policy, but also establishes a relation between the soft action-value $Q^s_\pi(a,s)$ and the policy $\pi$. To act according to $\pi$, we need a way to infer actions from Q-values. There are three main viable approaches. Firstly, we could learn a parametric Q-function and then project the exponentiated Q-values onto a parametric $\pi$. This approach has been considered in \citet{haarnoja2018soft} and was extended to use rich parametric policies in \citet{haarnoja2018latent}. Secondly, we could aim to sample from $\pi$ directly, for instance via importance sampling based on samples from $\pit(a|s)$ reweighed with $\exp(Q^s_{\pi}(a,s)/\alpha)$~\cite{haarnoja2017reinforcement}. This approach is known to have high-variance and is compute intensive. Finally, we could parameterise the Q-function, restricting its expressiveness, such that we can obtain $\pi$ in closed form. Using a Gaussian distribution for the policy would recover the NAF setting~\citep{gu2016continuous}, but this restricts $Q^s_{\pi}(a,s)$ to be quadratic in action space. 
In this paper we follow the third approach yet make use of a rich policy class of normalising flows, allowing the soft action-value function $Q^s_\pi$ to be a universal function approximator.

To achieve this we first solve Equation~\ref{eq:start} for $Q^s_{\pi}(a,s)$ to find the following equation.
\begin{equation}
    Q^s_{\pi}(a,s) = V^s_\pi(s) + \alpha \log \frac{\pi(a|s)}{\pit(a|s)} \quad \textrm{where} \quad V^s_\pi(s)=\alpha \log \int \pit(a|s)\exp(Q^s_\pi(a,s)/\alpha) \diff a
    \label{eq:Q}
\end{equation}
It makes sense to call the first term the soft value function, since taking the expectation of both sides of the equation gives $\EXP_\pi\big[ Q^s_{\pi}(a,s)\big] = V^s_\pi(s) + \alpha \KL[\pi,\pit|s]$, which corresponds to the definition of the soft value function. Additionally, the second term in Equation~\ref{eq:Q} can be interpreted as a soft version of the advantage function $A(a,s) = \alpha \log(\pi(a|s)/\pit(a|s))$, where differences in log-likelihoods are interpreted as advantages. Given this sum, a natural way to parameterise $Q$ becomes apparent. We can choose to parameterise $V^s_\pi(s)$ as a deep neural network, and $\pi(a|s)$ as a density modelled by a normalising flow~\cite{rezende2015variational}. These can be universal density estimators~\cite{huang2018neural} and hence allow $Q^s_\pi$ to model arbitrary functions. In the following, we chose to use a Real NVP architecture~\cite{dinh2016density} for our policy, as we can both sample and infer the probability density function efficiently\footnote{We note that to the best of our knowledge there is no formal proof that Real NVP's are universal density function approximators, nor any counterexamples of why they would not be. Other flows such as Neural Autoregressive Flows~\cite{huang2018neural} could be used when a formal proof is required.}. In order to condition the Real NVP on the state $s$, we concatenate $s$ to the input of every neural network inside the Real NVP. 

Using this parametrisation, we can fit $Q^s_{\pi}(a,s)$ directly by minimising the squared temporal difference error 
\begin{equation}
    \min_{\theta,\phi} \mathbb{E}_{\mu_\pi(s), p} \left[ \Big( r(s, a) + \gamma V^s_\pi(s'; \phi') - Q^s_{\pi}(a, s; \theta, \phi)  \Big)^2 \mid s' \sim p(s' | s, a)  \right],
\end{equation}
where $\theta$ denote policy parameters, $\phi$ are value function parameters and $\phi'$ are the parameters of a target value function, that are periodically copied from $\phi$; and $Q^s_{\pi}$, $V^s$ are given as in Equation \ref{eq:Q}. We approximate the expectation over transition and state visitation distribution by samples from a replay buffer. A full algorithm listing of the procedure is given in Algorithms \ref{alg:actor} and \ref{alg:critic}.


\begin{figure}[htpb]
\centering
\raisebox{0.02\textwidth}{\includegraphics[height=0.2\textwidth]{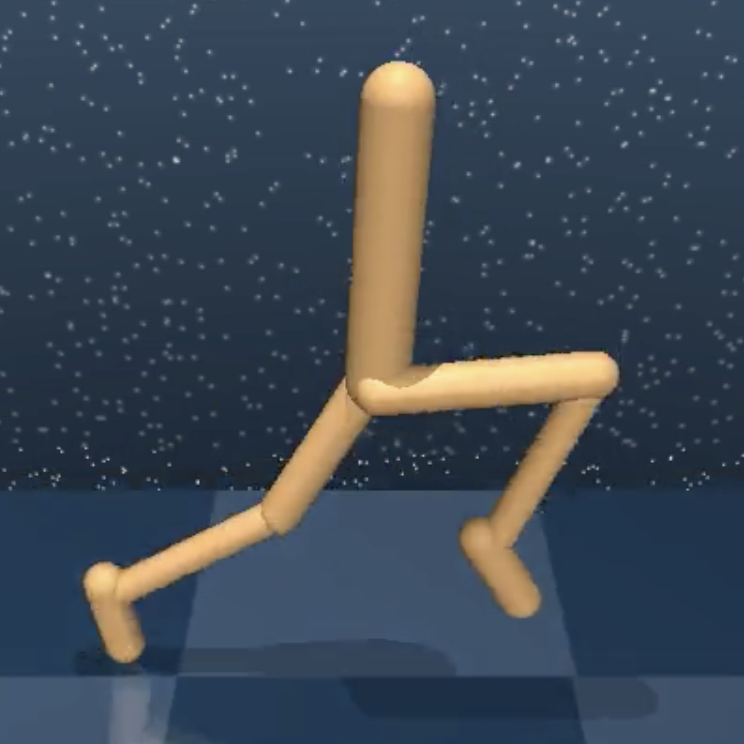}}
\raisebox{0.002\textwidth}{\includegraphics[height=0.249\textwidth]{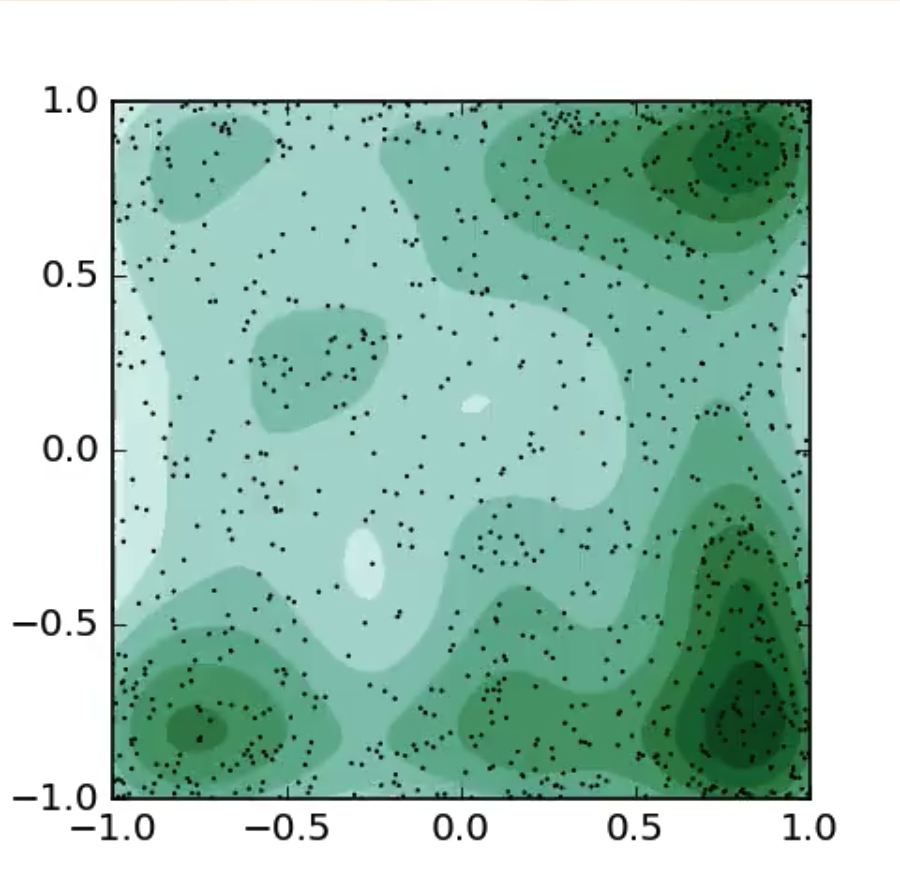}}
\raisebox{-0.002\textwidth}{\includegraphics[height=0.249\textwidth]{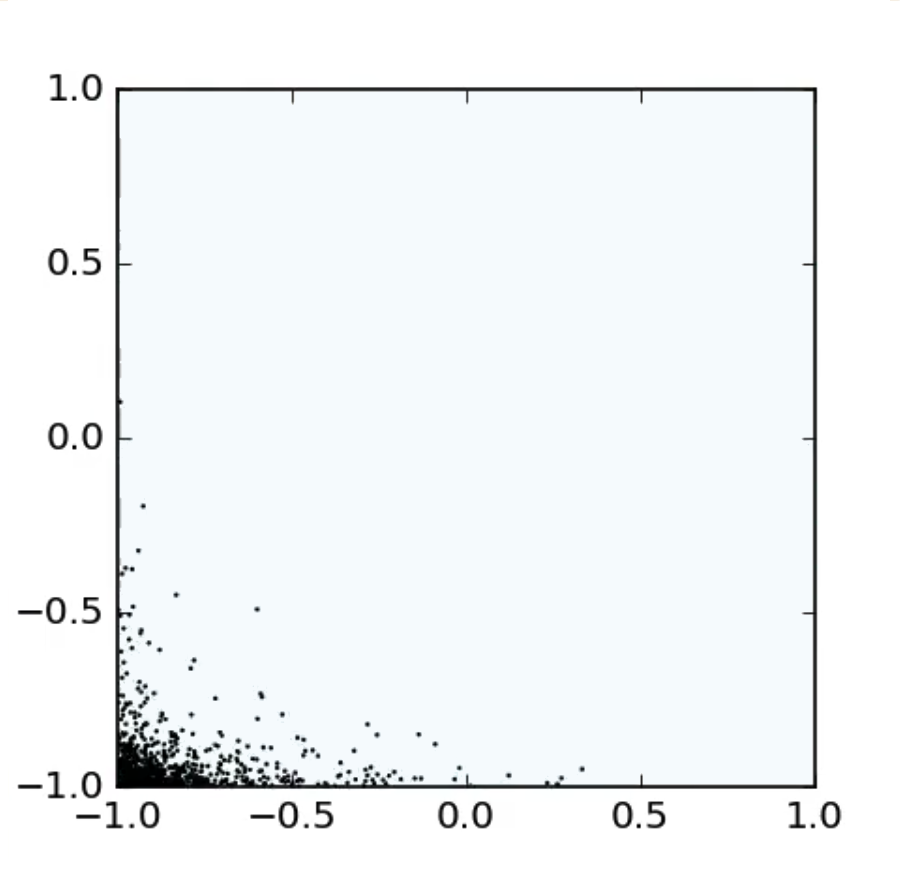}}
\includegraphics[height=0.249\textwidth]{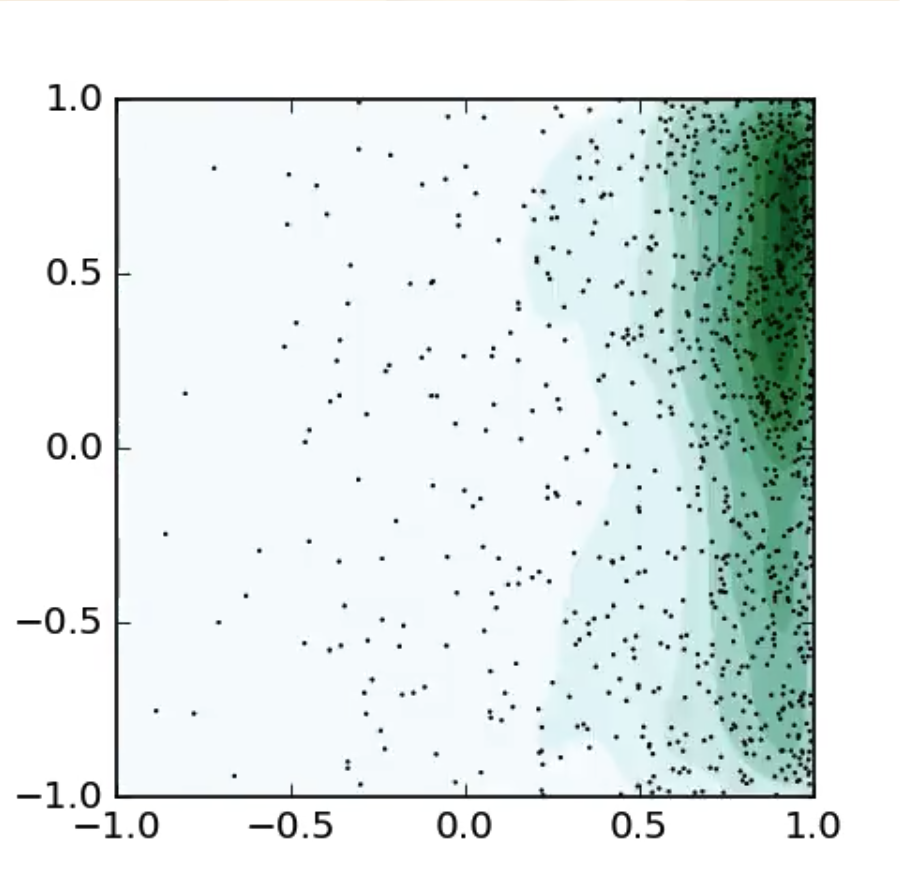}
\caption{An illustration of the distribution of the policy $\pi(a|s)$ in the state of the walker $s$ illustrated on the left. In the three scatter plots on the right we plotted 1000 randomly drawn $a\sim\pi(\cdot|s)$, where respectively action dimension $a_1$ is scattered against $a_2$, $a_3$ against $a_4$ and $a_5$ against $a_6$. To make the density differences clearer, we added a kernel density estimation on these samples. Note the finite support of the policy. As can be seen, some properties of the richer policy class are utilised, such as having high-skew and non-linearly correlated exploration noise. In the first scatter plot the policy also displays multimodal behaviour, with a mode in at least three corners of the action domain of the marginalised distribution. It is apparent that the resulting policy is not normally distributed.}
\label{fig:policy_sample}
\end{figure}

\section{Results}
We ran experiments across three domains from the DeepMind control suite~\cite{tassa2018deepmind}, the walker, the cheetah and the hopper, as depicted in Figure~\ref{fig:charts}. Our neural networks were initialised such that $Q(a,s)$ is identically zero in all states and actions, which means that our initial policy $\pi(a|s)$ is exactly uniform. All neural networks have weight normalisation with an initialisation based on the statistics of the first batch~\cite{salimans2016weight}. In order to deal with the gradients of the squashing operations in the Real NVP, we clip the gradient norm to $1$. We set the learning rate to $0.001$ and update the target network every $1000$ steps.

As we can see, the performance of Quinoa is similar to the one obtained by SVG(0)~\cite{heess2015learning} for the cheetah and the walker tasks. On the hopper task, the performance is slightly lacking behind.

When analysing the policies obtained, we find that using this richer class of distributions for a policy shows a distinct behaviour which is hard to obtain using a Gaussian policy. First of all, the actions samples from this policy have limited support. As shown in Figure~\ref{fig:policy_sample}, we can observe that in some states the policy has high-skew and non-linearly correlated exploration noise during the training process. Therefore, it is clearly not following a normal distribution. 

Moreover, the policy shows some multimodal behaviour during training. This can be explained by the fact that the converged policy for the walker domain has actions in the extremities for most states. The policy depicted has not converged yet, but it has learned that it prefers to take actions in the extremities. In this state however, it does not know which one yet, resulting in a multimodal distribution. The scatter plots also show how some dimensions of the action space have already collapsed, while others remain high in variance in order to keep the entropy large and keep exploring the action space.

\section{Conclusion}

In this paper, we describe a new parametrisation of the soft Q-function, such that the optimal policy can be obtained in closed form. This approach removes the need for a policy optimisation step from the learning process, simplifying standard actor-critic algorithms. We show that our algorithm is able to work across a range of tasks. We have illustrated that the policy is able to have an arbitrary distribution for its exploration noise. Moreover, we have shown that given this additional degree of freedom, the resulting policy does not show a Gaussian behaviour, with long tails and non-linearly correlated noise. We find in some states the actions of the policy are distributed multimodally. In the future, we will work on expanding this approach to harder tasks.

\bibliographystyle{abbrvnat}
\setlength{\bibsep}{5pt} 
\setlength{\bibhang}{0pt}
\bibliography{template_refs}

\section*{Citing this work}
This is a free, open access paper provided by DeepMind.
The final version of this work was published on workshops at NeurIPS on 2018-12-08. \textit{Cite as:} 
\citeas{degrave2018quinoa}

\section*{Funding}
This research was funded by DeepMind. The authors declare no competing financial interests.

\newpage

\section{Appendix}
\subsection{Solving the Q-function in closed form}
We want to find the optimal policy $\pi(a|s)$ which optimises $Q^s_\pi(a,s)$ subject to a hard constraint on the relative entropy between the policy $\pi$ and the prior $\pit$.
\begin{equation*}
\pi = \ARGMAX_\pi \EXP_{s\sim\mu_\pi} \big[Q^s_\pi(a,s)\pi(a|s)\big] \enspace\textrm{subject to}\enspace \EXP_{s\sim\mu_\pi}\big[\KL[\pi,\pit|s]\big] < \epsilon\enspace\textrm{and}\enspace \forall s: \EXP_{a}\big[\pi(a|s)\big] = 1,
\end{equation*}
where the last constraint ensures that $\pi$ is normalised. We solve this constrained optimisation problem using the method of Lagrange multipliers, obtaining an optimal $\alpha$ for a given $\epsilon$ automatically. We solve this constrained optimisation problem using the method of Lagrange multipliers. Here, the Lagrange function we construct is
\begin{equation*}
\LOSS = \EXP_{s\sim\mu_\pi} \big[Q^s_\pi(a,s)\pi(a|s)\big] + \alpha \big(\epsilon - \EXP_{s\sim\mu_\pi}\big[\KL[\pi,\pit|s]\big]\big) + \beta(s)\big(1 - \EXP_{a}\big[\pi(a|s)\big]\big)  
\end{equation*}
with Lagrange multipliers $\alpha \geq 0$ and $\beta(s)$. Next we maximise the Lagrangian $\LOSS$ w.r.t the primal variable $\pi$. The derivative w.r.t $\pi(a|s)$ for an action $a$ in a state $s$, making the approximation that $\mu_\pi(s)$ and $Q^s_\pi(a,s)$ are independent of $\pi$ is the following:
\begin{equation*}
\frac{\partial \LOSS}{\partial \pi} = Q^s_\pi(a,s) - \alpha \log\frac{\pi(a | s)}{\pit(a | s)} - \alpha - \beta(s).
\end{equation*}
Setting this derivative to zero, we find the policy which satisfies the Lagrangian in every state $s$.
\begin{equation*}
\pi(a|s) = \pit \exp(Q^s_\pi(a,s)/\alpha)\exp(- 1 - \beta(s)/\alpha)
\end{equation*}
Taking into account the constraint that $\pi(a|s)$ is a distribution, we obtain that the optimal policy $\pi(a|s)$ for a given $Q^s_\pi(a,s)$ is given by $\softmax_a(\log\pit(a|s) + Q^s_{\pi}(a,s)/\alpha)$\footnote{Here, we define a continuous softmax operation as $\softmax_x(y) = \frac{\exp(y)}{\int\exp(y) \diff x}$}, from which Quinoa derives its name. Note that $\pi(a|s)$ is always positive, so we have fulfilled the two conditions for it to be a probability density function.

At this point we can derive the dual function, by substituting the parametrisation for $Q^s_\pi(a,s)$ in the Lagrangian $\LOSS$.
\begin{equation*}
\LOSS(\alpha) = \alpha \epsilon + \alpha \EXP_{s\sim\mu_\pi}\Bigg[\log\EXP_{a\sim\pi}\Bigg[\exp\Big(\frac{V^s_\pi(s)}{\alpha} + \log \frac{\pi(a|s)}{\pit(a|s)}\Big) \Bigg]\Bigg]
\end{equation*}
When we minimise this convex function in $\alpha$, we find the optimal temperature for our distribution. Even though there is no analytic solution to this equation, the temperature can be computed efficiently using regula falsi to find the zero in the derivative under the constraint $\alpha \geq 0$:

\begin{equation*}
\frac{\partial \LOSS(\alpha)}{\partial \alpha} = \epsilon +  \EXP_{s\sim\mu_\pi}\Bigg[\log\EXP_{a\sim\pi}\Bigg[\exp\Big(\frac{Q^s_\pi(a,s)}{\alpha}\Big) \Bigg]
- \EXP_{a\sim\pi}\Bigg[\frac{Q^s_\pi(a,s)}{\alpha} \softmax_a\Big( \frac{Q^s_\pi(a,s)}{\alpha}\Big) \Bigg]\Bigg]
\end{equation*}

Finally, we have all the elements to write the algorithm for both the actor and the learner, which run asynchronously in parallel. These are written out in Algorithm~\ref{alg:actor} and Algorithm~\ref{alg:critic}.

\begin{algorithm}
  \caption{Actor algorithm}
  \label{alg:actor}
  \begin{algorithmic}[1]
    \REQUIRE policy $\pi(a|s)$ with parameters $\theta$ shared with the learner
    \REQUIRE replay buffer $\rho$ shared with the learner
    \REQUIRE environment $e$
    \ALOOP {while $\pi(a|s)$ not converged} 
        \ALOOP {while $e$ not terminated}
            \STATE get state $s$ from environment $e$
            \STATE sample $a$ from policy $\pi(\cdot | s)$
            \STATE send $a$ to environment $e$
        \ENDALOOP
        \STATE send trajectory to replay buffer $\rho$
    \ENDALOOP
  \end{algorithmic}
\end{algorithm}

\begin{algorithm}
  \caption{Learner algorithm}
  \label{alg:critic}
  \begin{algorithmic}[1]
    \REQUIRE policy $\pi(a|s)$ with parameters $\theta$ shared with the actor
    \REQUIRE replay buffer $\rho$ shared with the actor
    \REQUIRE prior policy $\pit(a|s)$ with parameters $\tilde{\theta}$
    \REQUIRE soft value $V_\pi^s(s)$ with parameters $\phi$
    \REQUIRE KL-constraint $\epsilon$
    \REQUIRE Discount $\gamma$
    \ALOOP {while $\pi(a|s)$ not converged} 
        \STATE sample $(s, a, r, s')$ from $\rho$
        \STATE $\KL = \log\pi(a|s) - \log\pit(a|s)$
        \STATE find optimal $\alpha$ minimising $\alpha \, \epsilon + \alpha \, \log \mathbb{E}_{a}[ \exp {(V_\pi^s(s)/\alpha + \KL)}]$
        \STATE $q = \alpha \KL + V_\pi^s(s)$
        \STATE $q' = r + \gamma V_\pi^s(s')$  and stop gradient
        \STATE optimise $\phi$ and $\theta$ to minimise $(q - q')^2$ with gradient descent
        \STATE every 1000 iterations: $\pit(a|s) \gets \pi(a|s)$
    \ENDALOOP
  \end{algorithmic}
\end{algorithm}

\end{document}